\PassOptionsToPackage{table,dvipsnames}{xcolor}
\documentclass[letterpaper]{article} 
\usepackage{aaai2026}  
\usepackage{times}  
\usepackage{helvet}  
\usepackage{courier}  
\usepackage[hyphens]{url}  
\usepackage{graphicx} 
\usepackage{natbib}  
\usepackage{caption} 
\usepackage{algorithm}
\usepackage{algorithmic}

\usepackage{enumitem}
\usepackage{natbib}

\usepackage{algorithm}
\usepackage{amssymb}       
\usepackage{multirow}                 
\usepackage{subfigure}

\usepackage[table]{xcolor}
\usepackage{xcolor}
\usepackage{microtype}
\usepackage{graphicx}

\usepackage{amsmath}
\usepackage{mathtools}
\usepackage{amsthm}

\usepackage{caption}
\usepackage{arydshln}

\usepackage{listings}
\usepackage{placeins}
\usepackage{nicefrac}
\usepackage{xurl}
\usepackage{enumitem}
\usepackage{tabularx}
\usepackage{nicefrac}
\usepackage{makecell}
\usepackage{xspace}
\usepackage{graphbox}
\usepackage{pifont}
\usepackage{enumitem}
\usepackage{colortbl}
\usepackage{array}
\usepackage{etoolbox}
\usepackage{footmisc}
\usepackage{dsfont}
\usepackage{afterpage}
\usepackage[textsize=tiny]{todonotes}

\newcolumntype{C}[1]{>{\centering\arraybackslash}p{#1}}
\newcolumntype{L}[1]{>{\raggedright\arraybackslash}p{#1}}
\newcolumntype{R}[1]{>{\raggedleft\arraybackslash}p{#1}}
\newcommand\Tstrut{\rule{0pt}{2.6ex}}         %
\newcommand\Bstrut{\rule[-0.9ex]{0pt}{0pt}}   %

\usepackage{newfloat}
\usepackage{listings}
\usepackage{cuted}

\usepackage[dvipsnames]{xcolor}
\usepackage{amsmath}
\usepackage{algorithmic}
\usepackage{algorithm}
\usepackage{array}
\usepackage{textcomp}
\usepackage{url}
\usepackage{verbatim}
\usepackage{xspace}
\usepackage{caption}
\usepackage{makecell}
\usepackage{booktabs}
\usepackage{lipsum}

\newtheorem{proposition}{Proposition}
\theoremstyle{definition}           
\newtheorem{assumption}{Assumption}

\usepackage{newfloat}
\usepackage{listings}
\DeclareCaptionStyle{ruled}{labelfont=normalfont,labelsep=colon,strut=off} 
\floatstyle{ruled}
\newfloat{listing}{tb}{lst}{}
\floatname{listing}{Listing}
\providecommand{\pdfinfo}[1]{}
\title{TIMA: Text-Image Mutual Awareness for Balancing Zero-Shot Adversarial Robustness and Generalization Ability}
\author{
    Fengji Ma\textsuperscript{\rm 1},
    Hei Victor Cheng\textsuperscript{\rm 2$^*$}, Chenxing Li\textsuperscript{\rm 3}, Li Liu\textsuperscript{\rm1 \thanks{Corresponding authors}}
}
\affiliations{
    \textsuperscript{\rm 1}Hong Kong University of Science and Technology (Guangzhou)\\
    \textsuperscript{\rm 2}Aarhus University\\
    \textsuperscript{\rm 3}Tencent AI Lab\\
    avrillliu@hkust-gz.edu.cn, hvc@ece.au.dk
}

\begin{document}

\maketitle


\begin{abstract}
Achieving zero-shot adversarial robustness without sacrificing generalization remains challenging for foundation models such as CLIP, especially under large adversarial perturbations. Through empirical analyses, we identify three critical yet overlooked issues: (1) Logit margins exhibit a stable offset between small and large adversarial perturbations, suggesting that explicitly adjusting margins could improve robustness against unseen large perturbations. (2) A significant negative correlation exists between logit margin and inter-class semantic similarity, indicating that semantic structures are insufficiently leveraged by existing methods. (3) Existing methods for adjusting text embeddings disrupt the intrinsic semantic consistency established by pre-trained models, undermining generalization capability. Motivated by these findings, we propose a novel {Text-Image Mutual Awareness (TIMA)} framework, including a {Text-Aware Image (TAI)} tuning module with an {Adaptive Semantic-Aware Margin (ASAM)} to explicitly calibrate logit margins, and an {Image-Aware Text (IAT)} tuning module with {Semantic Consistent Minimum Hyperspherical Energy (SC-MHE)} to preserve semantic consistency. Comprehensive experiments validate that TIMA significantly outperforms existing approaches by effectively addressing the identified limitations. 
\end{abstract}


\begin{figure}[ht!]
    \centering
    \includegraphics[width=\linewidth]{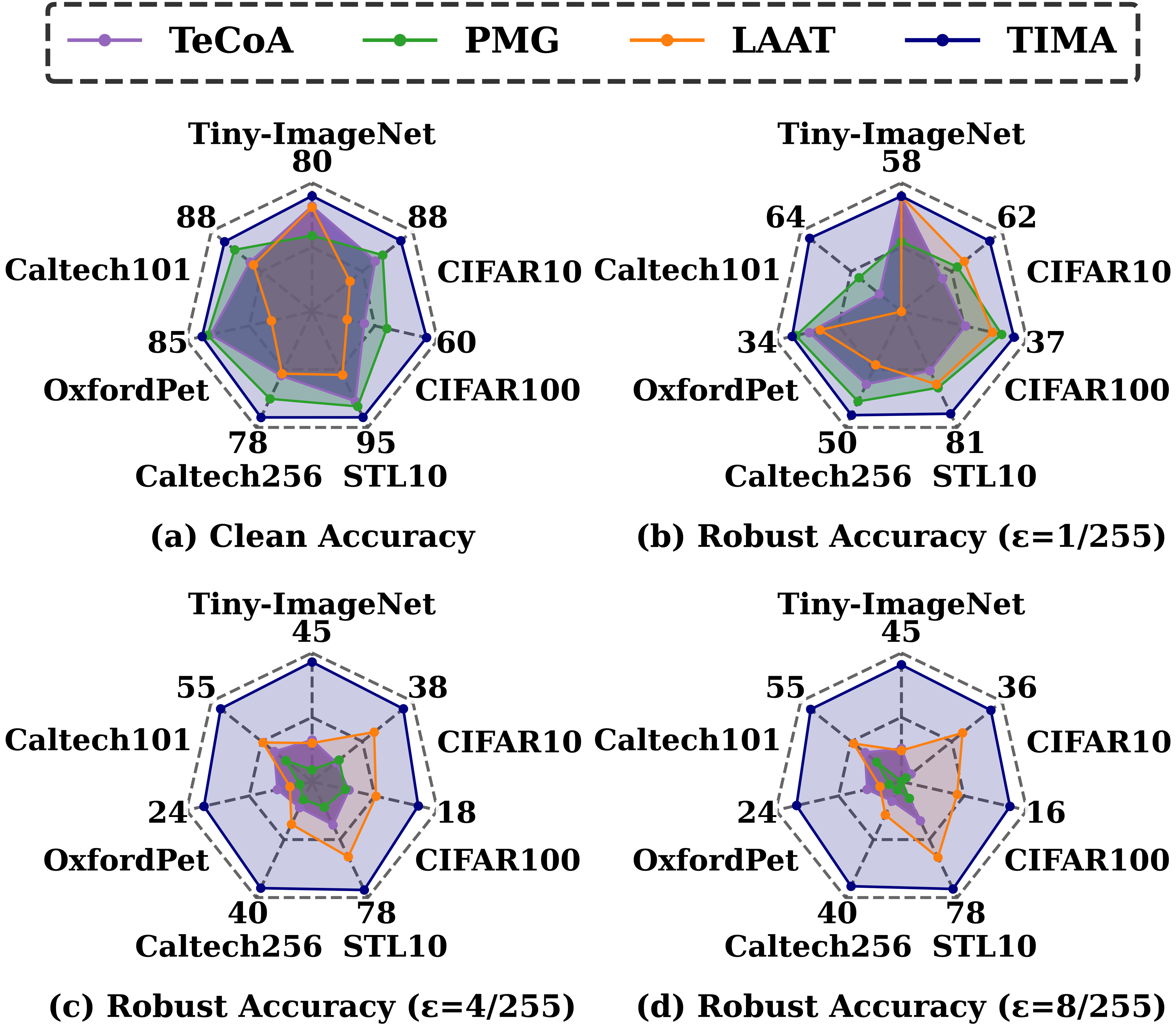}
    \caption{Zero-shot accuracies of compared methods and our proposed method. We show zero-shot robust accuracy of different methods under different perturbation radii. Better view by zooming in.}
    \label{fig:radar} 
\end{figure}

\section{Introduction}
\label{sec:intro}

Large-scale foundation models \cite{CLIP,foundations1,foundations2,foundations3} have recently garnered significant attention due to their excellent zero-shot generalization ability. Contrastive Language-Image Pre-training (CLIP)~\cite{CLIP}, the most widely used vision-language model, demonstrates the ability to accurately classify new classes with only simple text captions, even those that have not been encountered before. However, recent studies~\cite{TeCoA,LVL-AR,Robust-CLIP,CLIP-OOD-detecors,reading,ChatGPT-Robustness,Attack-foundation-vision,li2025towards} revealed that while these foundation models exhibit strong generalization performance, they are extremely vulnerable to adversarial perturbation. Notably, existing adversarial methods ~\cite{AT,CW,AA} have effectively attacked these foundation models, resulting in performance degradation of up to 90\% on different datasets, even with minor attacks. Meanwhile, recent works~\cite{li2025pbcat,li2025TIFS,li2024partimagenetpp} on robust recognition and detection further emphasize the practical importance of robustness in real-world systems: PartImageNet++~\cite{li2024partimagenetpp} scales up part-based models for robust recognition on ImageNet~\cite{ImageNet}, and adversarially robust object detectors are advanced by improving backbones~\cite{li2025TIFS} and employing patch-based composite adversarial training against physically realizable attacks~\cite{li2025pbcat}.

Traditionally, time-intensive and resource-heavy adversarial training is needed to achieve adversarial robustness. To bypass retraining across new tasks or datasets and to harvest the full potential of foundation models, the concept of \textbf{zero-shot adversarial robustness} in foundation models~\cite{TeCoA,LAAT,PMG} has emerged as a new and urgent research topic. The objective of zero-shot adversarial robustness is twofold: to {transfer adversarial robustness} in a zero-shot learning manner, and to {preserve zero-shot generalization} capabilities of foundational models. Here, adversarial robustness refers to a model's ability to withstand adversarial attacks with various perturbations. Zero-shot generalization, on the other hand, refers to the model's capacity to achieve high accuracy on previously unseen classes.

In pursuit of a zero-shot adversarial robust CLIP model, the pioneering work TeCoA~\cite{TeCoA} proposed aligning adversarial image embeddings with their corresponding text embeddings. This eliminates the need for retraining in subsequent visual tasks. Building on this idea, PMG~\cite{PMG} introduced additional constraints from pre-trained models and clean samples into the objective function, encouraging the preservation of generalized pre-trained information. Unlike the above two methods, which use only an adversarial fine-tuned image encoder, LAAT~\cite{LAAT} proposed an algorithm to expand the distance between fixed textual embeddings in hyperspherical space, showing promise against larger perturbation radii. However, existing methods fail to achieve zero-shot adversarial robustness while facing large perturbations (as shown in Fig.~\ref{fig:radar}) that are unseen during adversarial training.

To understand these limitations, our comprehensive empirical analyses (in Sec.~\ref{sec:motivation}) have identified three important insights: \textbf{1)} We observe a positive stable offset in logit margins when transitioning from small adversarial perturbations (e.g., $\varepsilon=1/255$) to a range of larger, unseen perturbations (such as $\varepsilon=2/255, 4/255,$ and $8/255$). Specifically, we find that the differences in logit margins under varying adversarial attack perturbation bounds can be approximated by a stable constant across different training epochs. This phenomenon indicates that explicitly calibrating logit margins during training could substantially enhance robustness against larger, unseen adversarial perturbations. \textbf{2)} We discover a significant negative correlation between the logit margin under adversarial perturbations and the semantic similarity between classes. Our analysis shows that categories with higher semantic similarity suffer from larger logit margin reductions, suggesting that existing methods fail to effectively leverage inherent semantic relationships to improve robustness. \textbf{3)} We identify that current methods aimed at increasing inter-class distances among text embeddings (e.g., LAAT~\cite{LAAT}) inadvertently disrupt the intrinsic semantic consistency originally learned by pre-trained CLIP models. Specifically, expansions in text embedding distances are found to significantly diminish semantic coherence, thereby undermining zero-shot generalization performance.

\begin{figure*}[t!]
\includegraphics[width=\linewidth]{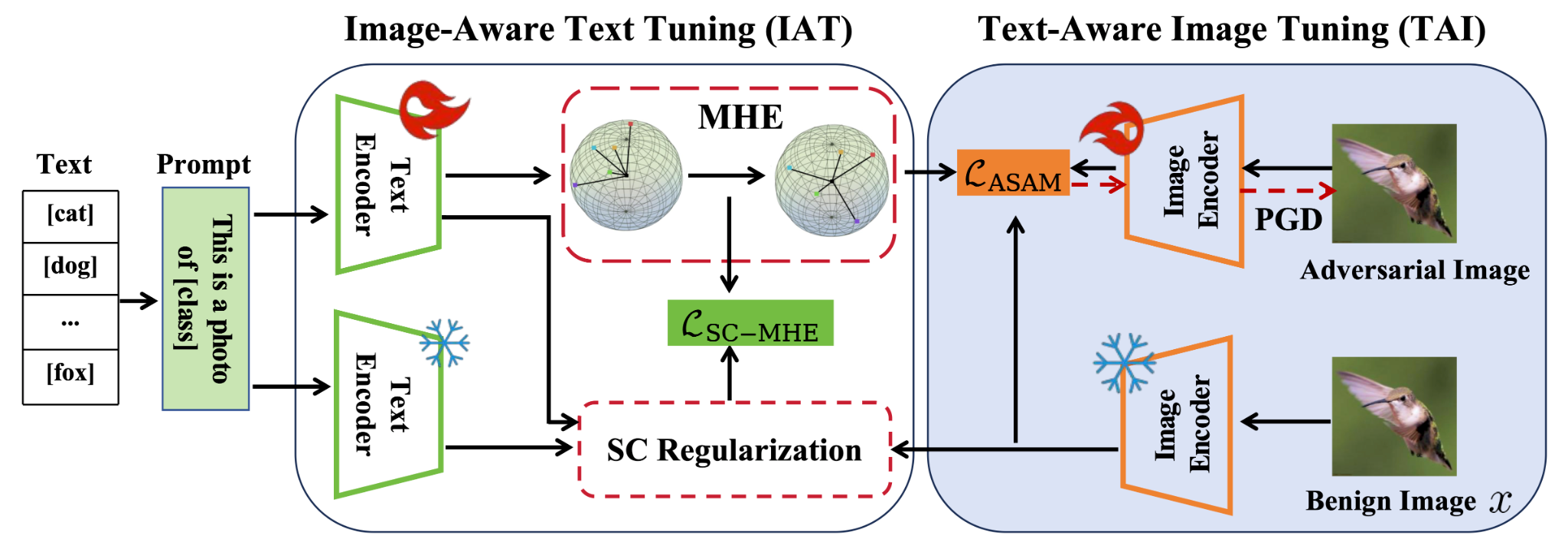}
\centering
\caption{Framework of our TIMA, including IAT and TAI tuning modules. SC-Regularization means semantic consistency regularization (Eq.(\ref{eq:kl-t})). $\mathcal{L}_{\mathrm{ASAM}}$ and $\mathcal{L}_{\mathrm{SC-MHE}}$ represent Adaptive Semantic-Aware Margin (ASAM) and Semantic Consistent Minimum Hyperspherical Energy (SC-MHE). Adversarial images are generated by Projected Gradient Descent (PGD).}
\label{fig:framework} 
\end{figure*}

Motivated by these insights, we propose a novel framework named \textit{Text-Image Mutual Awareness (TIMA)}, as shown in Fig.~\ref{fig:framework}, designed to simultaneously address these challenges. TIMA comprises two complementary modules: a) Text-Aware Image (TAI) tuning module: Motivated by the above insight 1) and 2), we introduce the \textit{Adaptive Semantic-Aware Margin (ASAM)}, explicitly calibrating decision boundaries according to class-level semantic similarity. By adaptively modulating logit margins during adversarial fine-tuning, TAI significantly improves robustness against large adversarial perturbations. b) Image-Aware Text (IAT) tuning module: Motivated by the above insight 3), we propose the \textit{Semantic Consistent Minimum Hyperspherical Energy (SC-MHE)} method. It uniformly enlarges inter-class distances among text embeddings on a hypersphere while employing semantic consistency regularization to preserve intrinsic semantic consistency from pre-trained embeddings, effectively maintaining zero-shot generalization capabilities.

The main contributions can be summarized as:
\begin{itemize}
\item This work proposes a novel TIMA framework, containing TAI and IAT. TIMA maximizes the contrastive and interactive information between text and image modalities, achieving a balance between zero-shot robustness and generalization, leading to effective multimodal training.
\item Motivated by insights from empirical analyses, by introducing an adaptive margin, we propose a new ASAM that provably increases inter-class distance and enhances zero-shot adversarial robustness. Notably, ASAM within the TAI tuning module is designed as a plug-and-play component, enabling seamless integration into any existing adversarial fine-tuning framework.
\item The proposed SC-MHE in the IAT tuning module achieves a trade-off between increasing text embedding inter-class distances and maintaining semantic information by semantic consistency regularization to prevent degradation in generalization performance. 
\item Extensive experiments show TIMA's superiority in zero-shot robust accuracy and clean accuracy in multiple datasets and under different perturbations, establishing its efficacy compared with state-of-the-art (SOTA) methods. Notably, TIMA demonstrates adversarial robustness to minor perturbation ($\varepsilon=1/255$) despite clean-data-only training.
\end{itemize}

\section{Related Works}
\label{sec:relatedworks}

\noindent\textbf{\textcolor{black}{Adversarial Robustness and Margin.}}
Prior works have sought to improve robustness by maximizing the margin from the decision boundary, using techniques like uniform loss objectives~\cite{MMA} or adaptive perturbation bounds per sample~\cite{CertifiedMargin,DynamicBoundary}. In contrast, our approach operates with a fixed training perturbation and instead improves robustness by identifying and enforcing a sample-level adaptive margin.

\noindent\textbf{\textcolor{black}{Zero-Shot Adversarial Robustness.}}
Existing methods for zero-shot adversarial robustness fall into three categories. The first, including TeCoA~\cite{TeCoA} and PMG~\cite{PMG}, fine-tunes the image encoder but struggles against large, unseen perturbations. The second, such as LAAT~\cite{LAAT}, adjusts text embeddings but can disrupt semantic consistency. A third category of training-free methods has recently emerged, which perform test-time defenses in either the feature~\cite{AOM} or pixel space~\cite{TTC}. In contrast, our TIMA framework introduces a Text-Image Mutual Awareness approach, co-optimizing image and text encoders via adaptive semantic-aware margins and semantic-consistent energy minimization to balance robustness and generalization.

\section{Methodology}

We begin by presenting the background and preliminaries on contrastive adversarial training and zero-shot adversarial robustness. Next, we describe the experimental observations that motivate our approach. Finally, we introduce the TIMA framework in detail, including its key components, as illustrated in Fig.~\ref{fig:framework}.

\subsection{Preliminaries}\label{sec:preliminaries}
We adopt the CLIP model as the foundation for our study, as it is a large-scale pre-trained vision-language model. The proposed method aims to train a foundational model that balances zero-shot robustness and generalization.  Given an image-text pair $(x, \mathcal{T})$, where $x$ represents an input image and $\mathcal{T}$ represents a text prompt with a fixed template ``This is a photo of \{\}", CLIP uses the image encoder $\phi$ and the text encoder $\psi$ encode both the image and the text in fixed-dimensional embeddings $z=\phi(x)$ and $t=\psi(\mathcal{T})$.

\noindent\textbf{Text-Guided Contrastive Adversarial Training.}
To generate the adversarial image $x^{\varepsilon}$ from a clean image $x$, TeCoA~\cite{TeCoA} employs the Projected Gradient Descent (PGD) method~\cite{AT}, aiming to find an $x^{\varepsilon}$ such that $||{x^{\varepsilon}-x||_\infty \le \varepsilon}$ which maximizes a chosen loss function (typically the cross-entropy loss w.r.t. the true label $y_c$). This is achieved through an iterative process involving gradient ascent and projection steps. The resulting adversarial image $x^{\varepsilon}$ is then used to compute the loss $\mathcal{L}$:
\begin{align}
    \mathcal{L}(x^{\varepsilon},\mathcal{T},y)= - \underset{i,j}{\sum}\bigg[y_{ij} \mathrm{log} \frac{\mathrm{exp}(\mathrm{s}(z^{\varepsilon}_{i},t_{j})/\tau)}{\underset{k}{\sum}\mathrm{exp}(\mathrm{s}(z^{\varepsilon}_{i},t_{k})/\tau)}\bigg], \label{eq:adv_loss} 
\end{align}
where $z^{\varepsilon}_{i} = \phi(x^{\varepsilon}_i)$ represents the adversarial image embedding of the $i$-th adversarial image example $x^{\varepsilon}_i$ encoded by the image encoder $\phi$. And $y_{ij} = 1$ if the image-text pair $(x_i, \mathcal{T}_j)$ is positive, otherwise, $y_{ij} = 0$. $\tau$ is the temperature parameter, and $s$ indicates calculating the cosine similarity of the two embeddings.

\noindent\textcolor{black}{\textbf{Logit Margin.}} We define the logit margin, a core concept for our analysis. For an adversarial image embedding ${z}^{\varepsilon}_i$ of an image ${x}_i$ with ground-truth class $c_i$, its logit margin $\gamma^{\varepsilon}_i$ is the difference between the cosine similarity with the most confusing class text embedding ${t}_k$ and the correct class embedding ${t}_{c_i}$:
\begin{align}
\gamma^{\varepsilon}_i &= \max_k\; \mathrm{s}({z}^{\varepsilon}_i, {t}_k) - \mathrm{s}({z}^{\varepsilon}_i, {t}_{c_i}).
\label{eq:logit-margin-adv}
\end{align}
The logit margin captures how adversarial perturbations shift image-text alignment away from the true class.
To quantify the margin's shift under stronger attacks, we further define the \textbf{perturbation-induced logit margin shift}. Specifically, we focus on the shift from a small training perturbation ($\varepsilon_{\text{small}}=1/255$) to a larger test-time perturbation $\varepsilon$:
\begin{align}
    \Delta\gamma_{i}^{(\varepsilon)} := \gamma^{(\varepsilon)}_{i}-\gamma^{(1/255)}_{i},
    \label{eq:margin_shift_definition}
\end{align}
which will be referred to throughout the rest of the paper as the perturbation-induced logit margin shift.

\begin{figure*}[t!]
\centering
\subfigure[\textcolor{black}{Perturbation-
Induced Logit Margin Shift ($\Delta\gamma_{i}$) \textit{vs.} Epoch}]
{
\begin{minipage}[b]{0.53\linewidth}
\includegraphics[width=1\linewidth]{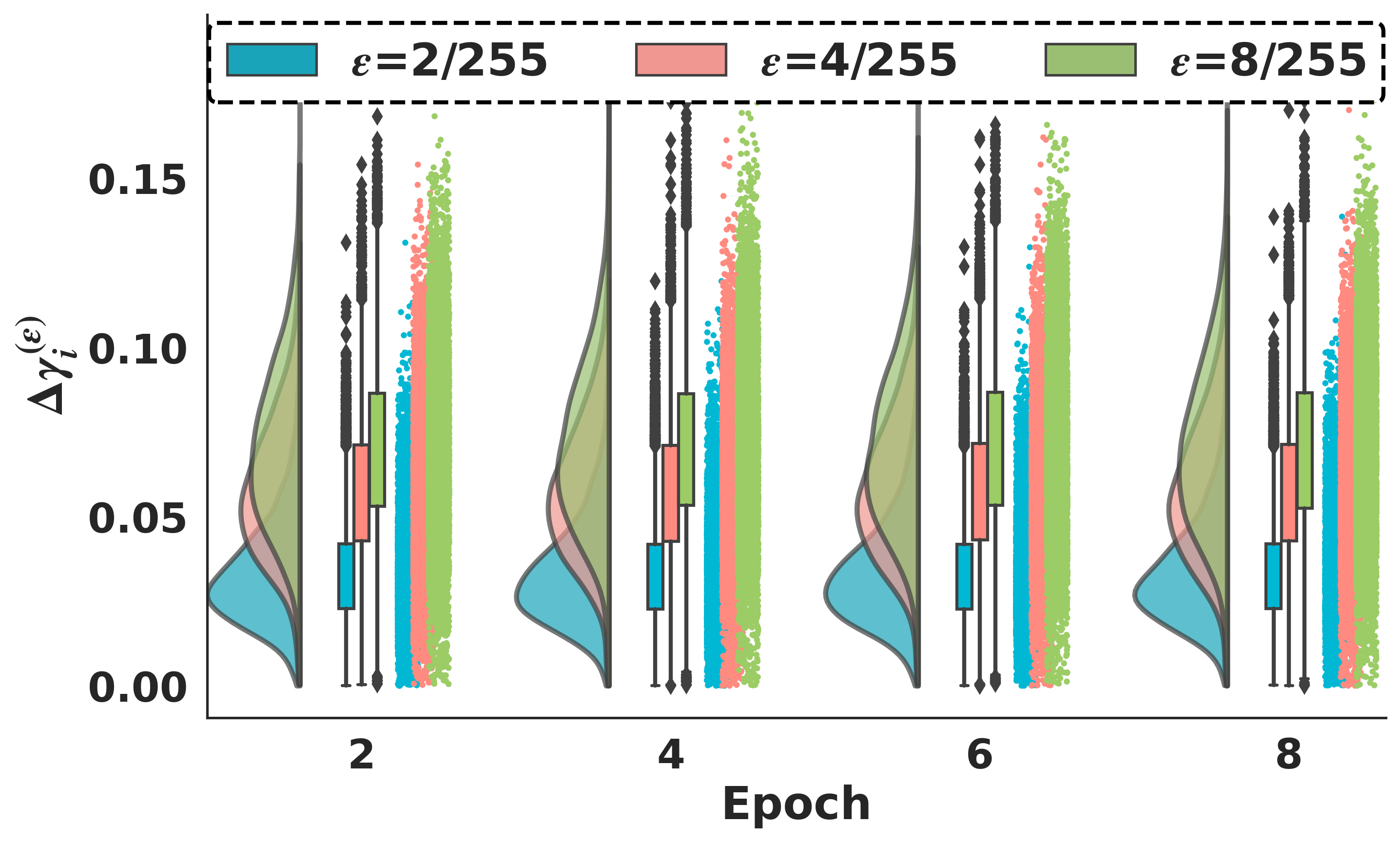}
\end{minipage}}
\subfigure[Logit Margin \textit{vs.} Semantic Similarity]
{
\begin{minipage}[b]{0.45\linewidth}
\includegraphics[width=\linewidth]{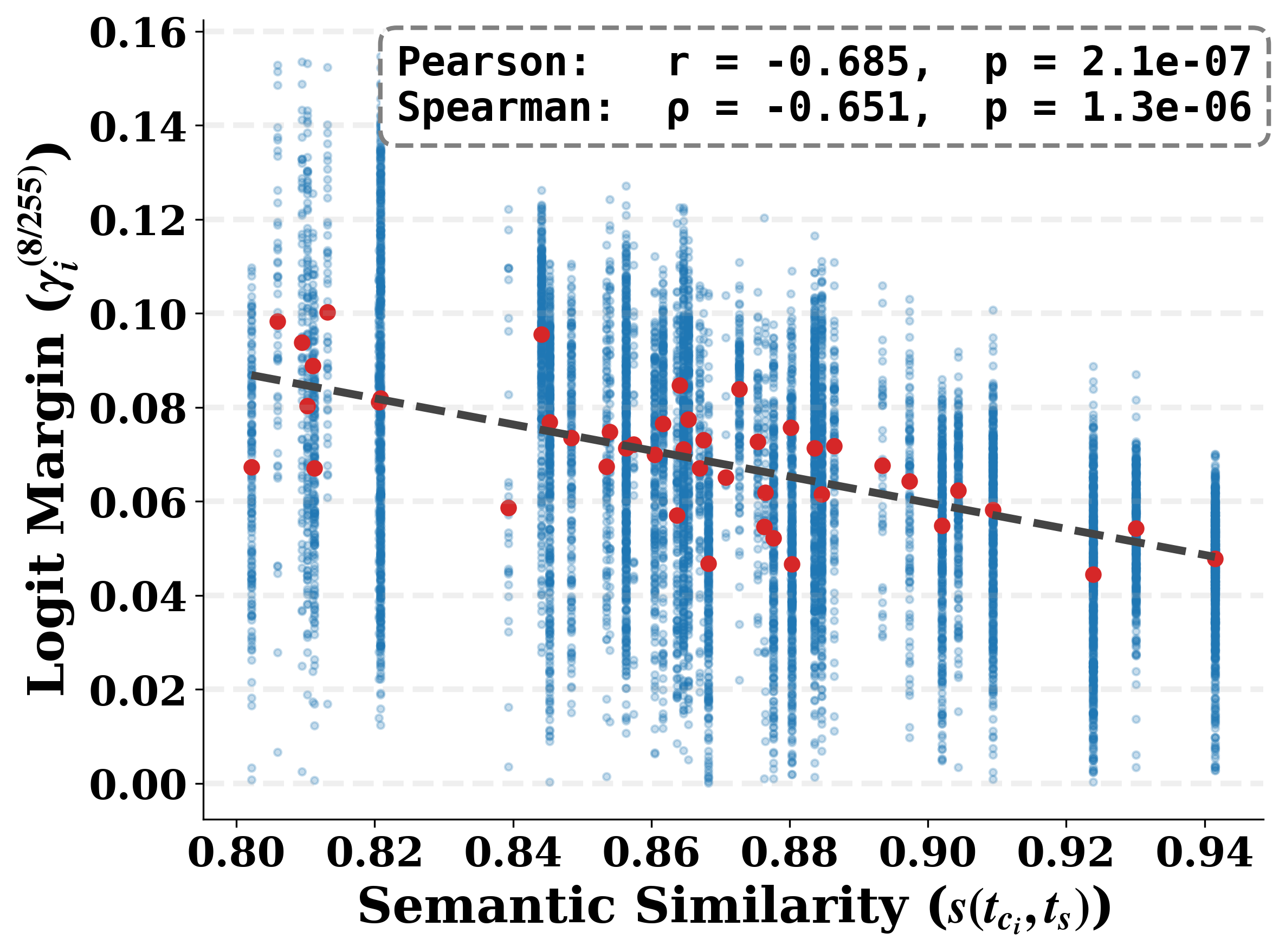}
\end{minipage}
}
\caption{Motivation for ASAM. (a) demonstrates the distribution of perturbation-induced logit margin shift $\Delta\gamma_{i}$ between large unseen ($\varepsilon=2,4,8/255$) and small seen ($\varepsilon=1/255$) perturbations, revealing a relatively stable offset during training. This stable margin offset indicates potential for improving robustness against unseen larger perturbations by adjusting decision boundaries during training. (b) illustrates the relationship between semantic similarity and logit margin under strong perturbation (\textit{e.g.} $\varepsilon = 8/255$). Each \textcolor{black}{blue dot} represents a sample \(x_i\), where the x-axis shows the semantic similarity between the ground-truth class \(c_i\) and the most confusing class \(s = \arg\max_k [s(z^{\varepsilon}_i, t_k) - s(z^{\varepsilon}_i, t_{c_i})]\), and the y-axis shows the corresponding logit margin \(\gamma_i^{\varepsilon} = s(z^{\varepsilon}_i, t_s) - s(z^{\varepsilon}_i, t_{c_i})\). \textcolor{black}{Red points} denote median logit margins aggregated across samples with similar semantic similarity.}
\label{fig:motivation}
\end{figure*}

\begin{figure}[t!]
\centering
\subfigure[\scriptsize CLIP]{
\begin{minipage}[b]{0.31\linewidth}
\includegraphics[width=1\linewidth]{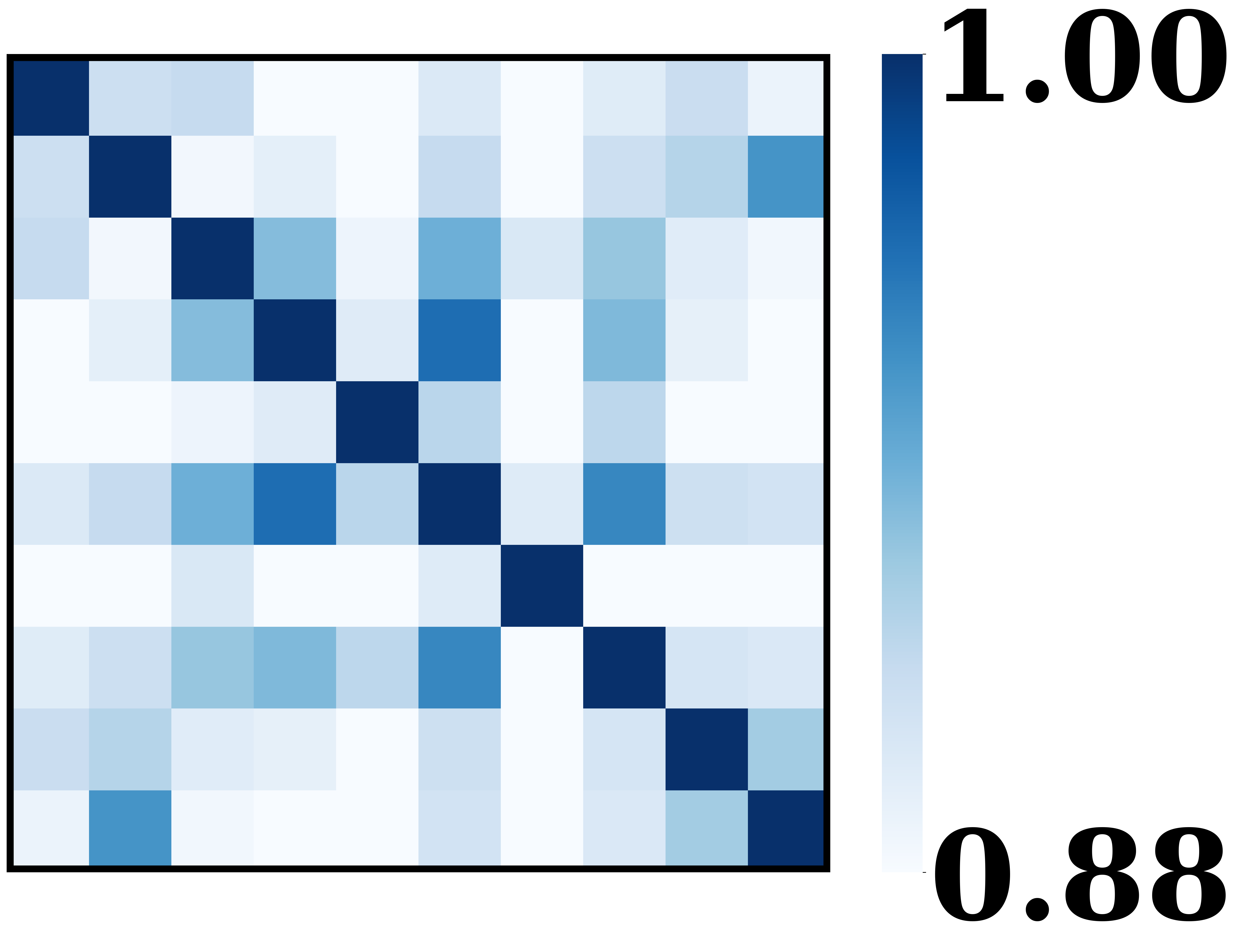}
\includegraphics[width=1\linewidth]{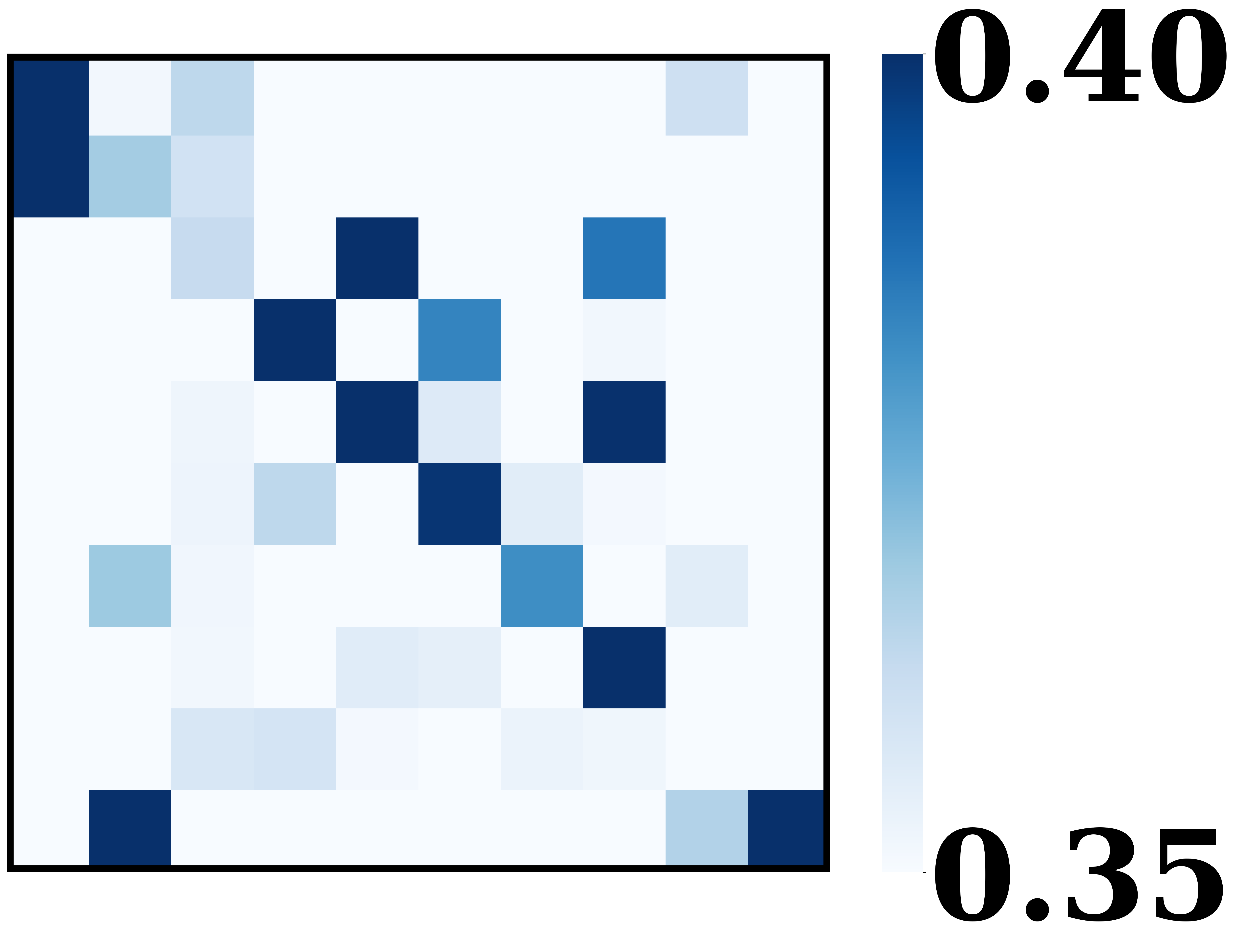}
\end{minipage}}
\subfigure[LAAT]{
\begin{minipage}[b]{0.31\linewidth}
\includegraphics[width=\linewidth]{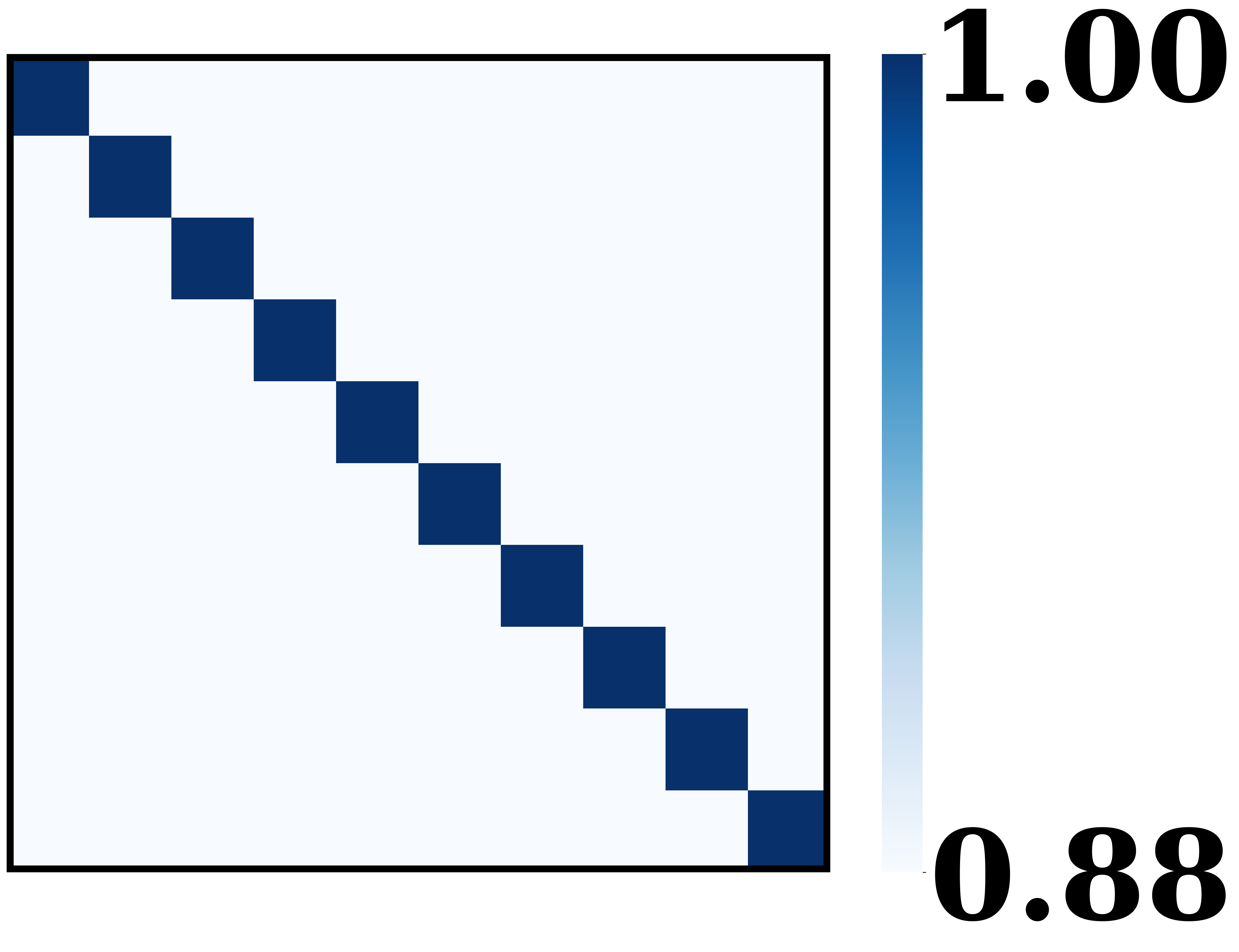}
\includegraphics[width=\linewidth]{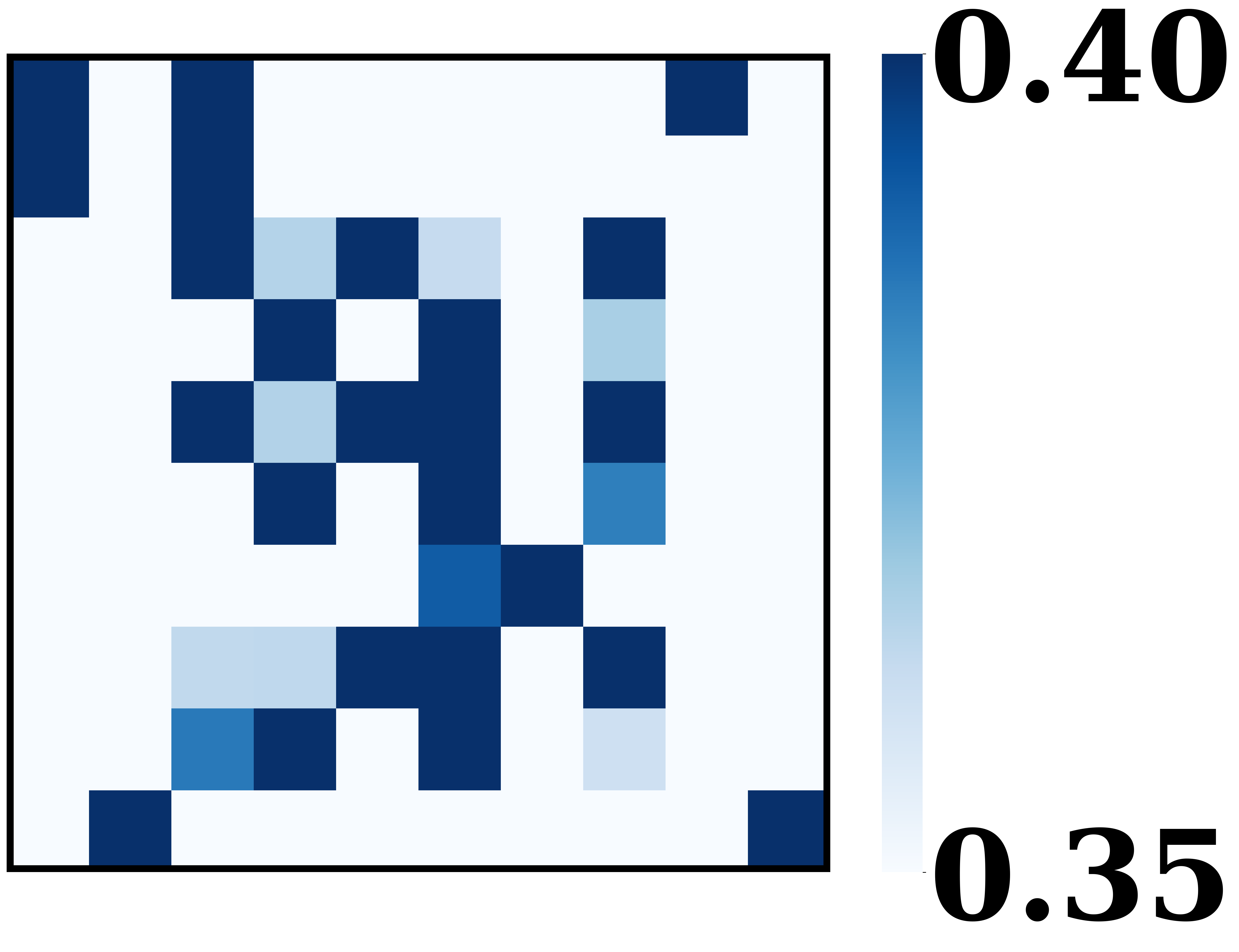}
\end{minipage}}
\subfigure[Ours]{
\begin{minipage}[b]{0.31\linewidth}
\includegraphics[width=\linewidth]{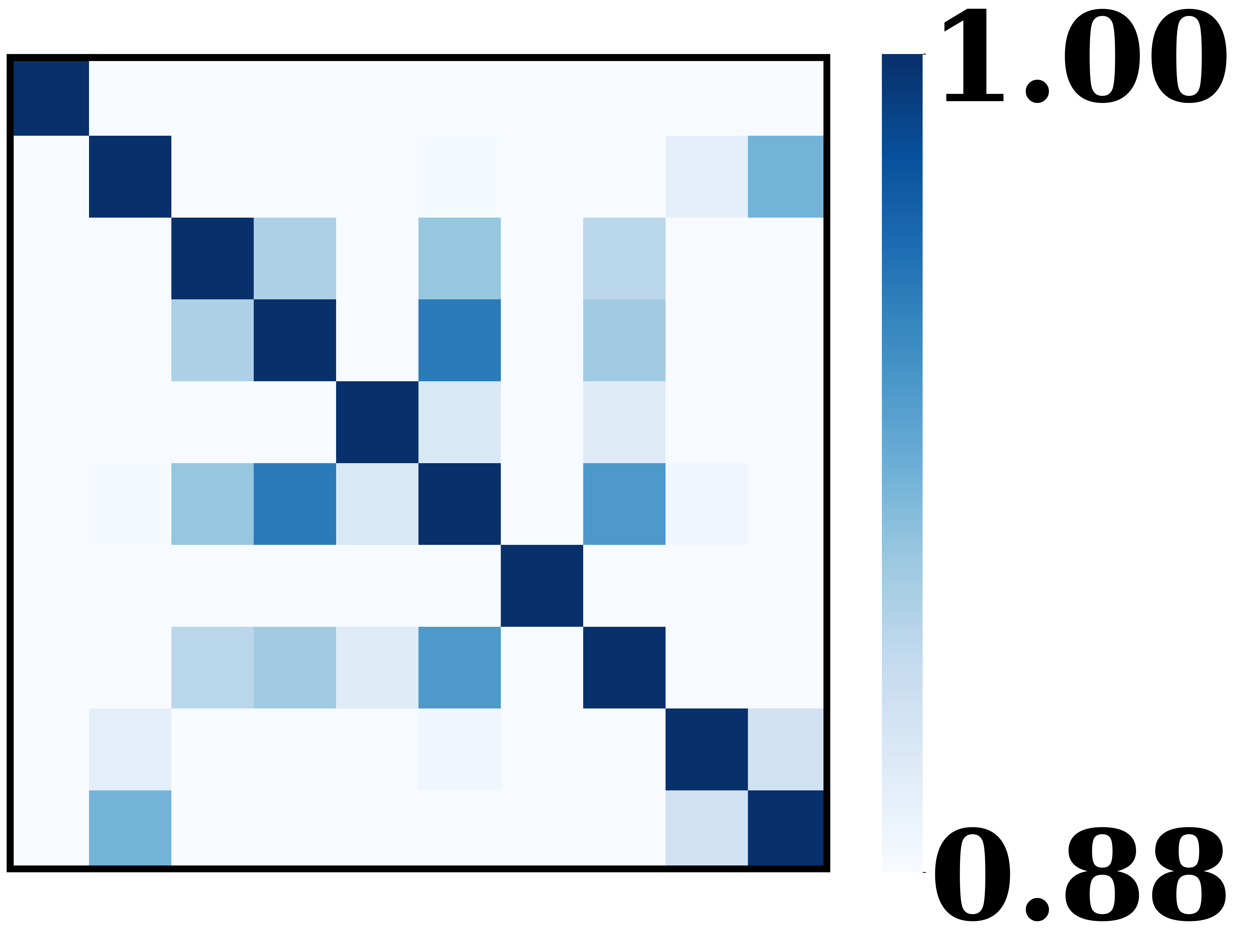}
\includegraphics[width=\linewidth]{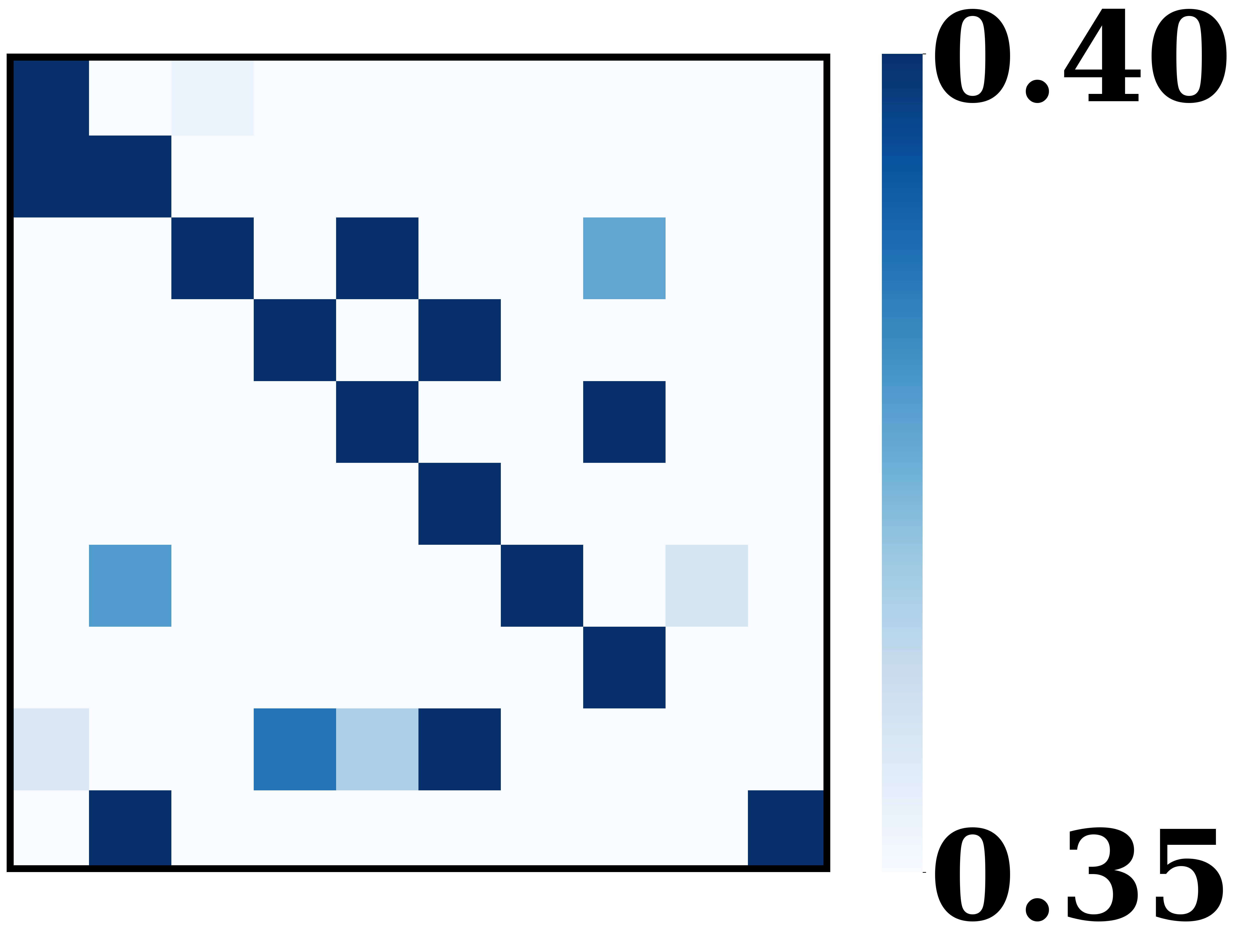}
\end{minipage}
}
\caption{Cosine similarity matrices on CIFAR-10. \textbf{Top row:} text-text similarity. \textbf{Bottom row:} clean image-text similarity. Our SC-MHE (c) better preserves the semantic structure of the original CLIP model (a) compared to LAAT (b).}
\label{fig:CLIP-LAAT-ours-text-text}
\end{figure}

\subsection{\textcolor{black}{Motivation}}
\label{sec:motivation}
To clarify the link between logit margin (Eq.(\ref{eq:logit-margin-adv}) and Eq.(\ref{eq:margin_shift_definition})) and robustness under large unseen perturbations, we analyze the behavior of the SOTA method TeCoA, focusing on logit margin shift, semantic preservation, and robustness.  We also examine why LAAT, despite its robustness, shows poor generalization.  These insights motivate our proposed approach.
\subsubsection{Perturbation-Induced Logit Margin Shift Analysis.}
We analyze the logit margin shift \(\Delta\gamma_i\) caused by perturbations, which measures how the model's class separation changes from small training-time perturbations (\(\varepsilon=1/255\)) to larger ones at test time (\(\varepsilon=2,4,8/255\)). The results show that \(\Delta\gamma_i\) remains stable across training epochs, suggesting a consistent positive gap between margins under large and small perturbations. This indicates that reducing margins under small perturbations can help improve robustness to larger ones. In addition, some samples show large \(\Delta\gamma_i\) values throughout training, meaning they are more sensitive to perturbation strength. These may be unclear or rare cases, highlighting the need for a sample-wise adaptive margin.
\begin{assumption}
\label{assumption1}
The image encoder $\phi$ is $L$-Lipschitz, implying its Jacobian's spectral norm is bounded: $\sup_{x' \in \mathcal{N}} \|J_\phi(x')\|_2 \le L$.
\end{assumption}
\begin{assumption}
\label{assumption2}
The Jacobian of the encoder $J_\phi$ is $M$-Lipschitz continuous: $\forall x_1, x_2 \in \mathcal{N}$, $\|J_\phi(x_1) - J_\phi(x_2)\|_2 \le M \|x_1 - x_2\|_2$.
\end{assumption}
\begin{proposition}
Based on Assumption~\ref{assumption1} and~\ref{assumption2}, there exists $K \le 2M$ such that $\|\nabla \gamma(x)\| \le 2L$ and, 
for any $0 < \varepsilon_1 < \varepsilon_2$,
\[
\bigl|\Delta\gamma - (\varepsilon_2 - \varepsilon_1)\,\|\nabla\gamma(x)\|\bigr|
\le \frac{K}{2}\,(\varepsilon_2^2 - \varepsilon_1^2).
\]
\end{proposition}

So the logit margin shift $\Delta\gamma$ is radius-ordered and deviation-bounded. As the bound grows, $\Delta\gamma$ increases in a stable, bounded-variation manner—exactly the trend in Fig.3(a). Full proof is in the \textit{Appendix}.

\subsubsection{Correlation Between Logit Margin and Semantic Similarity.}
\label{sec:margin-semantic}
In adversarially trained models (e.g., TeCoA), we further study the relationship between class-wise text embedding similarity and logit margins under strong perturbations. Fig.~\ref{fig:motivation}(b) plots the cosine similarity between the ground-truth class embedding \(t_{c_i}\) and its most similar class \(t_s\) (x-axis), and the corresponding logit margin \(\gamma^{(8/255)}\) under \(\varepsilon=8/255\) (y-axis). Each blue point is a sample, and red points are class medians. We observe a strong negative correlation (Pearson \(r=-0.685\), Spearman \(\rho=-0.651\), both \(p \ll 0.001\)), with the medians showing an approximately linear trend. This motivates our margin in Eq.(\ref{eq:margin-ori}).

Based on observations from Fig.~\ref{fig:motivation} (a) and (b), our \textbf{ASAM} design follows \textbf{three principles}: (1) adjusting margins under small perturbations improves robustness to larger ones; (2) the margin should reflect class-level semantic similarity; (3) it should adapt to individual samples.
\subsubsection{Semantic Inconsistency in Text Embedding Adjustment.}
While prior methods that adjust text embeddings, such as LAAT~\cite{LAAT}, successfully increase inter-class distances, they disrupt the intrinsic semantic consistency of the pre-trained CLIP model. As illustrated in Fig.~\ref{fig:CLIP-LAAT-ours-text-text}(b), this damages the inherent semantic relationships between similar classes (e.g., cats and dogs, cars and trucks), thereby degrading zero-shot generalization performance on clean data and motivating our proposed semantic-aware regularization.
\subsection{Text-Image Mutual Awareness}
\label{sec:TIMA}
\begin{table*}[!t]
\centering
\small
\tabcolsep=2.7pt
\extrarowheight=-0.8pt
\begin{tabular}{C{5mm} C{25mm}  C{6.5mm}   C{6.5mm} C{6.5mm} *{3}{C{6.5mm} C{6.5mm} C{6.5mm}} C{6.5mm} C{6.5mm}  C{6.5mm} C{1.5mm}}
\toprule
\multirow{2}{*}[-3.5em]{Eval.} & \multirow{2}{*}[-2.9em]{\makecell{Compared\\ Methods}} & & \multicolumn{13}{c }{Zero-shot datasets} & \multirow{2}{*}{}
\\  
\cline{4-16} 
 &  
 & \parbox[c][1.6cm]{0.8cm}{\centering{\rotatebox[origin=c]{90}{TinyIN}}}  
 & \parbox[c][1.2cm]{0.8cm}{\centering\rotatebox[origin=c]{90}{CIFAR10}} 
 & \parbox[c][1.2cm]{0.8cm}{\centering\rotatebox[origin=c]{90}{CIFAR100}} 
 & \parbox[c][1.2cm]{0.8cm}{\centering\rotatebox[origin=c]{90}{STL-10}} 
 & \parbox[c][1.2cm]{0.8cm}{\centering\rotatebox[origin=c]{90}{ImageNet}} 
 & \parbox[c][1.2cm]{0.8cm}{\centering\rotatebox[origin=c]{90}{OxfordPets}} 
 & \parbox[c][1.2cm]{0.8cm}{\centering\rotatebox[origin=c]{90}{Food}} 
 & \parbox[c][1.2cm]{0.8cm}{\centering\rotatebox[origin=c]{90}{SUN}} 
 & \parbox[c][1.2cm]{0.8cm}{\centering\rotatebox[origin=c]{90}{DTD}} 
 & \parbox[c][1.8cm]{0.8cm}{\centering{\rotatebox[origin=c]{90}{CalTech101}}} 
 & \parbox[c][1.8cm]{0.8cm}{\centering{\rotatebox[origin=c]{90}{CalTech256}}} 
 & \parbox[c][1.2cm]{0.8cm}{\centering\rotatebox[origin=c]{90}{StandfordCars}} 
 & \parbox[c][1.2cm]{0.8cm}{\centering\rotatebox[origin=c]{90}{FGVC}} 
 & \parbox[c][1.2cm]{0.8cm}{\centering\rotatebox[origin=c]{90}{Flowers}}
 & 
 \parbox[c][1.6cm]{0.8cm}{\centering{\rotatebox[origin=c]{90}{Average (\%)}}} 
\\
\toprule
\midrule
\parbox[t]{3mm}{\multirow{5}{*}{\rotatebox[origin=c]{90}{clean}}}
& CLIP &   60.74   & 89.06  & 62.32  & 97.16 & 59.12       & 85.82     & 83.15      & 57.67 & 40.52 & 85.47 & 82.02&52.17 & 8.40& 65.04&$\;$67.76 \\
\cdashline{2-18}  \Tstrut \Bstrut
& TeCoA  	 & 76.00	    & 76.10	   & 46.00	    & 91.39	& 43.87       & 75.33	  & 43.11	     &  46.15	 & 29.73 & 78.03& 69.08 &30.71 &7.03 & 40.82& $\;$53.81\\
& PMG   	& 70.98 
   & 78.47	   & 50.31	    & 92.10	 & 46.08	     & 76.62	        & 53.45		  & 45.85	    & 31.12 & {81.98} & 73.07 &35.79 &5.08 &46.29 &$\;$56.23  \\
& LAAT  	& 75.84
    & 68.08	   & 42.72	   & 87.75	 & 42.72	        & 51.21	   & 34.81	          & 43.92	& 18.14 & 77.12 & 68.73 &21.06 & 7.20& 29.68& $\;$47.78  &
\\
\cdashline{2-18}  
\Tstrut \Bstrut
& TIMA (Ours)   & \textbf{77.75	}   & \textbf{84.23	}	   & \textbf{57.90	}	    & \textbf{93.61}	& \textbf{46.55	}		       & \textbf{78.80}	     & \textbf{55.82}	  & \textbf{53.37	}	    & \textbf{32.55	}	 & \textbf{84.59}  & \textbf{76.27}  & \textbf{37.37}  &  \textbf{9.18} & \textbf{42.38} &   $\;$\textbf{59.31} \\
\midrule
\parbox[t]{3mm}{\multirow{5}{*}{\rotatebox[origin=c]{90}{$\varepsilon=\nicefrac{1}{255}$}}}
& CLIP  & 2.72     & 9.57    & 4.55     & 35.40 & 3.50         & 2.72      & 3.95        & 1.02  & 2.50  & 28.99& 20.27 & 0.66 &0.00 & 1.37&$\;$8.37  & 
\\
\cdashline{2-18} \Tstrut \Bstrut
& TeCoA     & 56.63	& 54.91	       & 30.15    & 75.14	& 12.98 	        & 29.76	   & 9.59	     	  & 13.86	 & 10.77 &45.26 & 35.13 & 3.95&0.00 & 13.82& $\;$28.00 & \\
& PMG     & 52.52	
    & 56.67	  & {34.19}	   & 76.89	 & \textbf{15.17}	        & 31.45	  & \textbf{13.10}	       &\textbf{16.30}	 & {10.82} & 50.11& 41.03 &5.03 &0.00 & 15.24& $\;$29.89 & \\
& LAAT     	& {56.72}   & 57.50   & 33.15	    & 76.54	& 13.30 	     & 28.35	        & 12.83	   	  & 13.53	 & 8.51  &32.90 & 28.42 &0.71 & \textbf{0.66}& 1.56& $\;$26.05 & \\
\cdashline{2-18}  
\Tstrut \Bstrut
& TIMA (Ours) &   \textbf{56.75	}	    & \textbf{60.53}		&\textbf{35.56	}	  & \textbf{79.58	}		  & {15.16}	       &\textbf{31.92}	    & {13.06}	 & 	        15.66	  & \textbf{14.36}  &   \textbf{61.88} &  \textbf{45.77}  & \textbf{6.75}  & 0.59	& \textbf{20.12}  &$\;$\textbf{32.69}
\\
\midrule
\parbox[t]{3mm}{\multirow{5}{*}{\rotatebox[origin=c]{90}{$\varepsilon=\nicefrac{8}{255}$}}}
& CLIP &  1.99 &  5.03 & 0.42 & 31.43  & 1.11& 4.91 & 10.10     &  0.36 &  0.05 &  22.23 & 13.62& 0.36  &  0.00 &  0.36  &  $\;$6.57  & \\
\cdashline{2-18} \Tstrut \Bstrut
& TeCoA     	& 15.29    & 7.06	   & 3.95	    & 40.93	 & 1.85	        & 6.54	     & 1.45	   & 1.10	 & 0.53 & 23.11 &15.03 &0.12 &0.00 &0.00 & $\;$8.35  &  \\
& PMG     	 & 4.13   & 5.51	   & 2.07	    & 30.06	& 0.54 	        & 2.34	    & 1.13	    & 0.42	  & 0.37 & 17.26 &12.13 &0.03 & 0.00& 0.00& $\;$5.43 & \\
& LAAT     	 & 14.67	  & 23.44	   & 9.34	   & 58.59	 & 1.04    & 4.11	          & 2.67	     & 1.62	 & 0.32 &28.73 & 18.61&0.22 & \textbf{0.12}&0.36  & $\;$11.70 &\\
\cdashline{2-18}  
\Tstrut \Bstrut
&TIMA (Ours)& \textbf{41.31}	   & \textbf{32.45}	   & \textbf{14.38}	   & \textbf{73.84}	 & \textbf{10.21}  	      & \textbf{20.05}	    &\textbf{9.55}	    &   \textbf{7.48} &\textbf{2.29}	   &\textbf{50.04} &  \textbf{37.06} & \textbf{1.18}  &  	0.00	 & \textbf{3.32}   & $\;$\textbf{21.65}  
\\
\bottomrule
\end{tabular} 
\caption{Clean and adversarial evaluation of CLIP model under PGD attack. The best accuracies are bold.}
\label{tab:zero-shot-tiny}
\end{table*}

\subsubsection{Adaptive Semantic-Aware Margin.}
\label{sec:AW}

Based on the above analysis, we introduce ASAM to explicitly adjust logit margins. The decision boundary in TeCoA's objective (Eq.(\ref{eq:adv_loss})) is $s({z}^{\varepsilon}_{i}, {t}_{k}) = s({z}^{\varepsilon}_{i}, {t}_{c_{i}})$. To improve robustness, we enforce a semantic-aware negative margin:
\begin{align}
s({z}^{\varepsilon}_{i}, {t}_{k}) - s({z}^{\varepsilon}_{i}, {t}_{c_{i}}) = - m \cdot (1 - s({t}_{k}, {t}_{c_{i}})),
\label{eq:margin-ori}
\end{align}
where $m > 0$ is a scaling factor. This constraint imposes a larger separation for more semantically similar classes, directly addressing the vulnerability observed in Fig.~\ref{fig:motivation}(b).

Furthermore, to achieve sample-wise adaptivity, we make the margin conditional on the sample's inherent ambiguity. Using a frozen image encoder $\phi_{\mathrm{frozen}}$ to get the clean embedding $\hat{{z}}_i = \phi_{\mathrm{frozen}}({x}_i)$, we define the adaptive margin $\mathcal{M}_{ik}$ as:
\begin{align}
    \mathcal{M}_{ik} = \begin{cases}
m \cdot \mathrm{s}({t}_{c_i}, {t}_k), & \text{if } \mathrm{s}(\hat{{z}}_i, {t}_k) \ge \eta \cdot \mathrm{s}(\hat{{z}}_i, {t}_{c_i}) \\
0, & \text{otherwise}
\end{cases}
\end{align}
where $\eta$ is a threshold. This mechanism selectively penalizes only the samples that are inherently easy to confuse.

Finally, we integrate this adaptive margin into the contrastive loss function to obtain the final ASAM loss:
\begin{align}
   \mathcal{L}_{\mathrm{ASAM}} = -\sum_{i,j} y_{ij} \log \frac{\exp([\mathrm{s}({z}^{\varepsilon}_i, {t}_j) - \mathcal{M}_{ij}]/\tau)}{\sum_k \exp([\mathrm{s}({z}^{\varepsilon}_i, {t}_k) - \mathcal{M}_{ik}]/\tau)},
\label{eq:TAI}
\end{align}
where $y_{ij}$ is the one-hot label. 

\subsubsection{Semantic Consistent Minimum Hyperspherical Energy.}
\label{sec:MHE}

To address the semantic degradation caused by prior text embedding adjustments (\textit{e.g.,} LAAT), we propose the {Semantic Consistent Minimum Hyperspherical Energy (SC-MHE)} method as the core of our {IAT} tuning module.

SC-MHE aims to uniformly distribute text embeddings on a hypersphere to maximize inter-class separation, while simultaneously preserving the semantic structure learned by the pre-trained model. This is achieved by a loss function with two components:
1) MHE, which encourages uniform spacing by maximizing the Euclidean distance $d(\cdot, \cdot)$ between text embeddings ${t}$.
2) Semantic Consistency Regularization, which prevents semantic drift during optimization. We introduce Semantic Consistency Regularization $\mathcal{R}$ to align the predictive distribution of the tuned text embeddings (${t}$) with that of the frozen, original embeddings ($\hat{{t}}$) on clean images as follows:
\begin{align}
\mathcal{R}_{j} = &\underset{i}{\sum}p^{tr}_{ij}\mathrm{log}(\frac{p^{tr}_{ij}}{p^{st}_{ij}}),
     \label{eq:kl-t}
 \end{align}
 where $p^{st}_{ij}=\frac{\mathrm{exp}(\mathrm{s}(\hat{z}_{i},t_{j})/\tau)}{\underset{k}{\sum}\mathrm{exp}(\mathrm{s}(\hat{z}_{i},t_{k})/\tau)}$ and $p^{tr}_{ij}=\frac{\mathrm{exp}(\mathrm{s}(\hat{z}_{i},\hat{t}_{j})/\tau)}{\underset{k}{\sum}\mathrm{exp}(\mathrm{s}(\hat{z}_{i},\hat{t}_{k})/\tau)}$. The combined SC-MHE loss function is:
\begin{equation}
\resizebox{0.9\linewidth}{!}{$
\mathcal{L}_{\mathrm{SC\text{-}MHE}} = \underset{j}{\sum} \left[ \underset{k \neq j}{\sum} \frac{1}{1 + d^{\alpha}(t_{j}, t_{k})} + \lambda_{\mathrm{T}} \underset{i}{\sum} p^{tr}_{ij} \log \frac{p^{tr}_{ij}}{p^{st}_{ij}} \right]
$}
\label{eq:IAT}
\end{equation}
where $\alpha=2$ and $\lambda_{\mathrm{T}}{=}1$. Due to space limits, we justify {why MHE improves zero-shot robustness} in {\textit{Appendix}}.

\begin{figure*}[htbp!]
\centering
\subfigure[TeCoA]{
\begin{minipage}[b]{0.225\linewidth}
\includegraphics[width=1\linewidth]{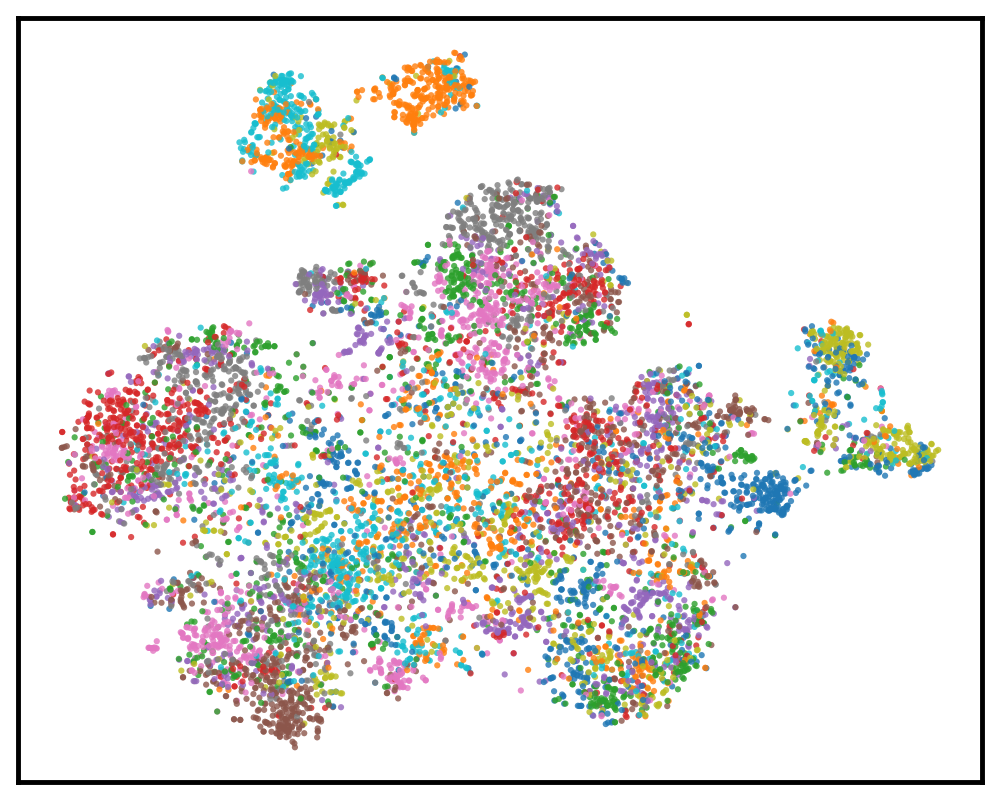}
\end{minipage}}
\subfigure[PMG]{
\begin{minipage}[b]{0.225\linewidth}
\includegraphics[width=1\linewidth]{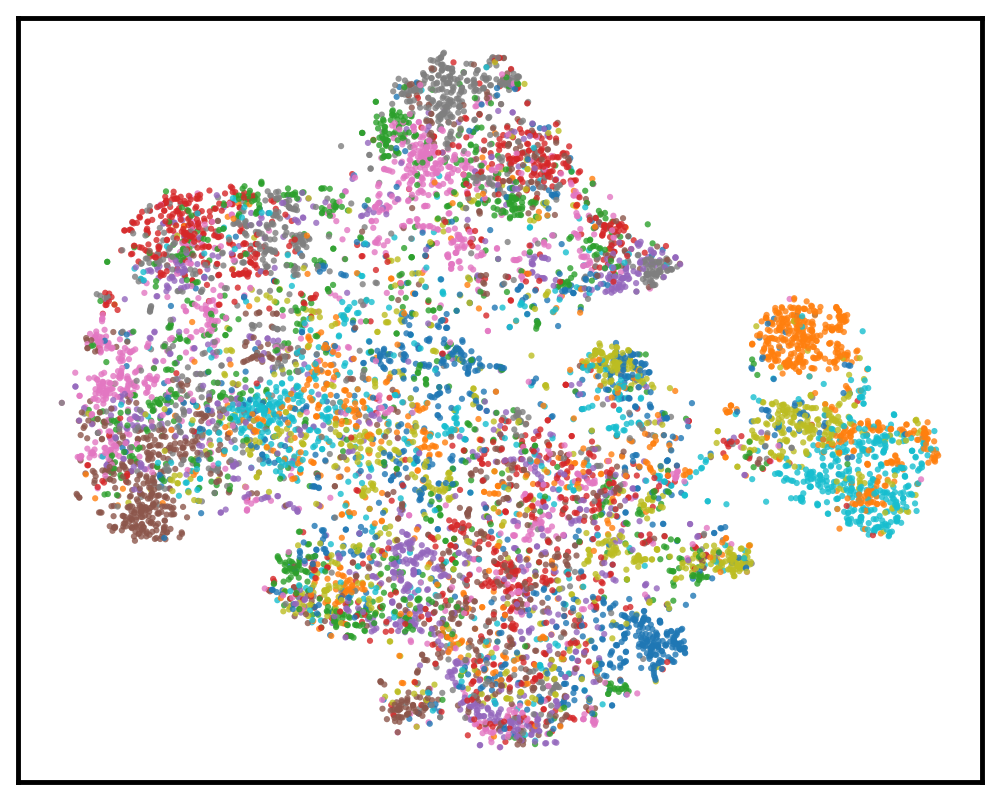}
\end{minipage}}
\subfigure[LAAT]{
\begin{minipage}[b]{0.225\linewidth}
\includegraphics[width=1\linewidth]{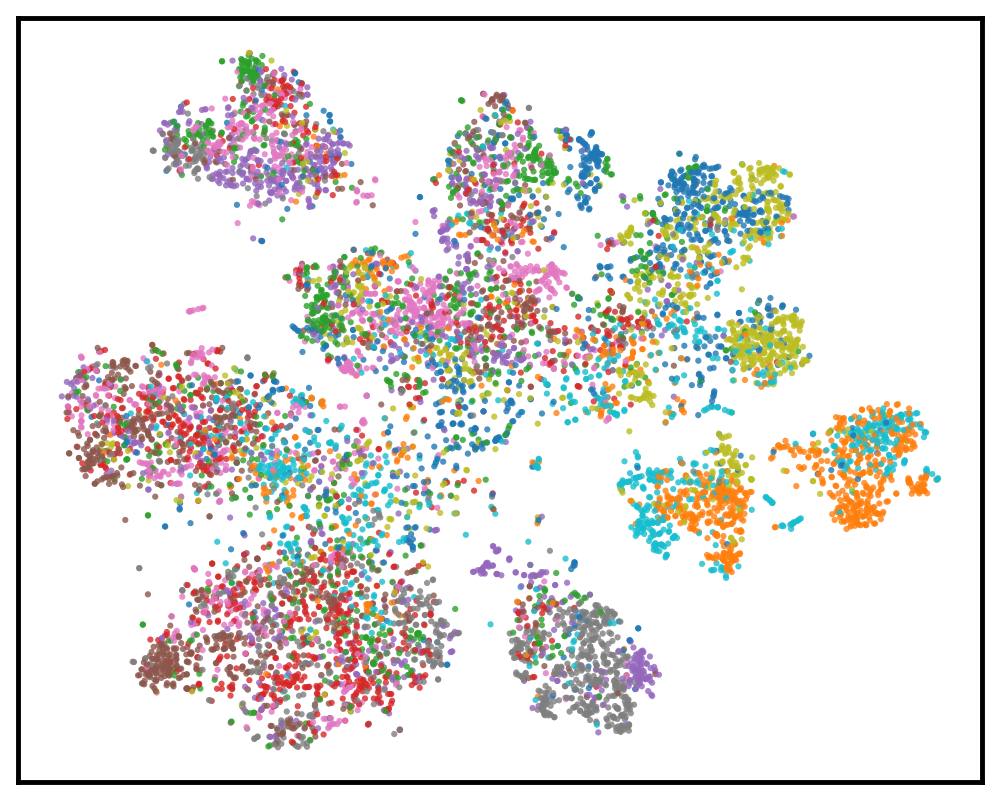}
\end{minipage}}
\subfigure[TIMA (Ours)]{
\begin{minipage}[b]{0.225\linewidth}
\includegraphics[width=1\linewidth]{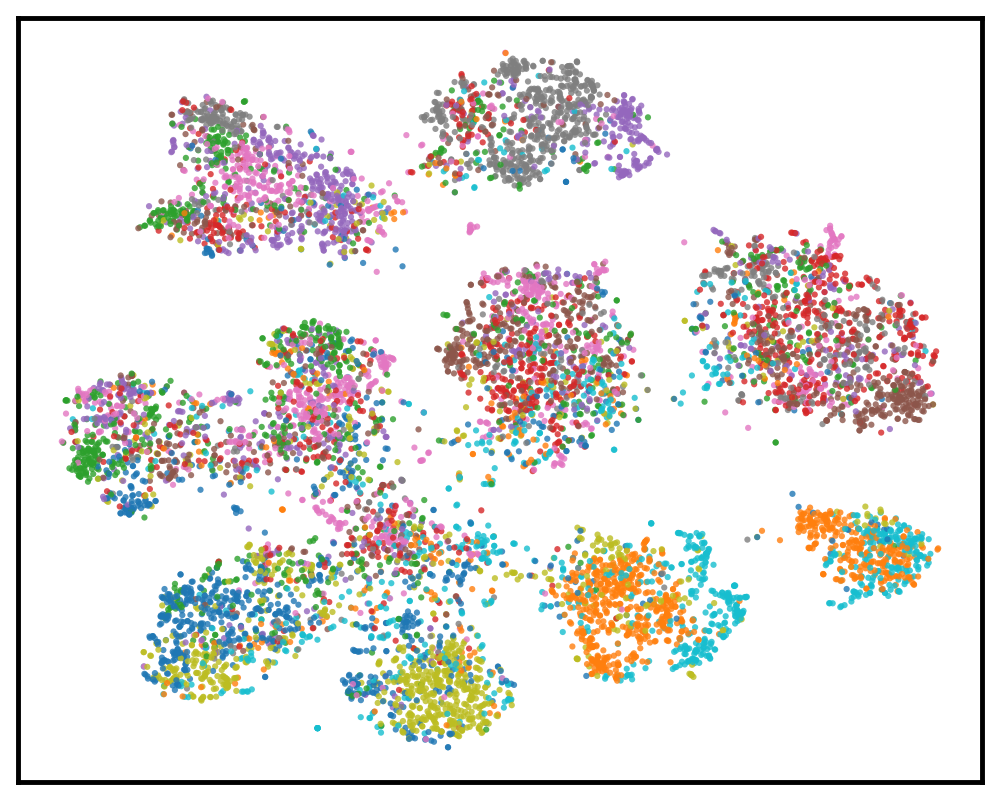}
\end{minipage}}
\caption{T-SNE visualizations of image embeddings on the CIFAR-10 dataset under large perturbation.}
\label{fig:tSNE}
\end{figure*}

\begin{table*}[htbp!]
\centering
\small
\tabcolsep=2.7pt
\extrarowheight=-0.8pt
\begin{tabular}{C{5mm} C{25mm}  C{6.5mm}   C{6.5mm} C{6.5mm} *{3}{C{6.5mm} C{6.5mm} C{6.5mm}} C{6.5mm} C{6.5mm}  C{6.5mm} C{1.5mm}}
\toprule
\multirow{2}{*}[-3.5em]{$\varepsilon$} & \multirow{2}{*}[-2.9em]{\makecell{Compared\\ Methods}} & & \multicolumn{13}{c }{Zero-shot datasets} & \multirow{2}{*}{}
\\  
\cline{4-16} 
 &  
 & \parbox[c][1.6cm]{0.8cm}{\centering{\rotatebox[origin=c]{90}{TinyIN}}}  
 & \parbox[c][1.2cm]{0.8cm}{\centering\rotatebox[origin=c]{90}{CIFAR10}} 
 & \parbox[c][1.2cm]{0.8cm}{\centering\rotatebox[origin=c]{90}{CIFAR100}} 
 & \parbox[c][1.2cm]{0.8cm}{\centering\rotatebox[origin=c]{90}{STL-10}} 
 & \parbox[c][1.2cm]{0.8cm}{\centering\rotatebox[origin=c]{90}{ImageNet}} 
 & \parbox[c][1.2cm]{0.8cm}{\centering\rotatebox[origin=c]{90}{OxfordPets}} 
 & \parbox[c][1.2cm]{0.8cm}{\centering\rotatebox[origin=c]{90}{Food}} 
 & \parbox[c][1.2cm]{0.8cm}{\centering\rotatebox[origin=c]{90}{SUN}} 
 & \parbox[c][1.2cm]{0.8cm}{\centering\rotatebox[origin=c]{90}{DTD}} 
 & \parbox[c][1.8cm]{0.8cm}{\centering{\rotatebox[origin=c]{90}{CalTech101}}} 
 & \parbox[c][1.8cm]{0.8cm}{\centering{\rotatebox[origin=c]{90}{CalTech256}}} 
 & \parbox[c][1.2cm]{0.8cm}{\centering\rotatebox[origin=c]{90}{StandfordCars}} 
 & \parbox[c][1.2cm]{0.8cm}{\centering\rotatebox[origin=c]{90}{FGVC}} 
 & \parbox[c][1.2cm]{0.8cm}{\centering\rotatebox[origin=c]{90}{Flowers}}
 & \parbox[c][1.6cm]{0.8cm}{\centering{\rotatebox[origin=c]{90}{Average (\%)}}}  
\\
\toprule
\midrule
\parbox[t]{3mm}{\multirow{2}{*}{\rotatebox[origin=c]{0}{0}}}
& TeCoA  	 &  76.61	 &  88.77		   &  	65.02	    &  96.31		&  51.06	     &  82.36	  &  	71.63	    &   57.83		 & 36.01	  &  84.77	&  79.98	& 45.16	  & 10.05	 & 46.34 & $\;$63.71 \\
& TIMA (Ours)   & 78.65	   & 90.41		   & 67.40		    & 96.08		& 54.23	    & 84.74	  & 75.61   & 61.90	 & 37.66	  & 85.80	  & 81.90	  & 47.02	  &  13.12	 & 48.71 &   $\;${65.95} \\
\midrule
\parbox[t]{3mm}{\multirow{2}{*}{\rotatebox[origin=c]{0}{$\nicefrac{1}{255}$}}}
& TeCoA     	&0.98 & 3.71	 & 1.76	    & 5.08		 &  0.04	 & 0.00	 &  0.00	 & 0.09	 & 0.00	 & 0.00	 & 0.33	 & 0.00	 &0.00	 & 0.13 & $\;$0.87  &  \\
&TIMA (Ours)& 16.43		   & 26.30	   & 12.66	   & 37.70	 & 4.74	   & 10.33	   & 10.18	    &   3.09	& 2.49	 	   & 19.57	&  18.53	 & 1.16	 &  	0.13	 	 & 0.78    & $\;${11.72}  
\\
\bottomrule
\end{tabular} 
\caption{Clean and adversarial evaluation under PGD attack on various datasets \emph{without adversarial training}.}
\label{tab:zero-shot-clean-training}
\end{table*}

\begin{figure}[ht!]
    \centering \includegraphics[width=\linewidth]{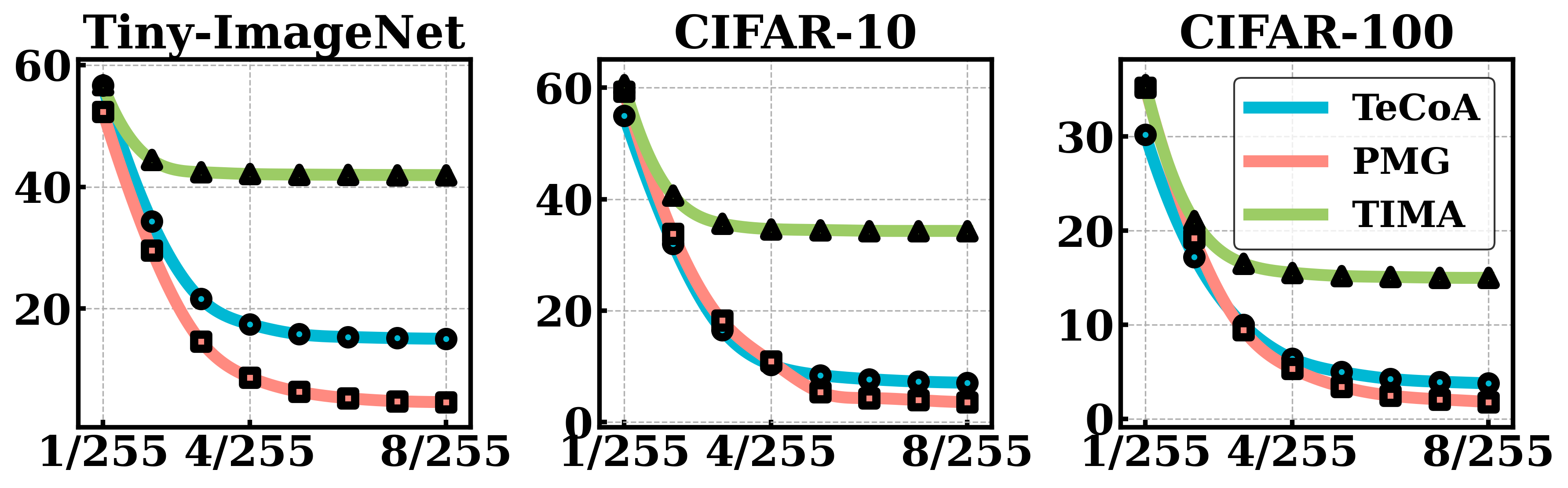}
    \caption{\textcolor{black}{{Adversarial evaluation under different perturbation bounds across different datasets at inference time.}
}}
    \label{fig:CW-AA-BIM}
\end{figure}

\section{Experiments}
\subsection{Experimental Setup}
\label{exp:datasets}
\noindent{\textbf{Dataset and Baseline.}} 
To evaluate zero-shot adversarial robustness and generalization, we consider 14 datasets spanning multiple recognition tasks, including CIFAR10, CIFAR100~\cite{CIFAR100}, STL10~\cite{STL10}, Caltech101~\cite{Caltech101}, Caltech256~\cite{Caltech256} for generic classification; OxfordPets~\cite{OxfordPets}, Food101~\cite{Food101}, Flowers~\cite{Flowers102}, StandfordCars~\cite{StandfordsCar}, FGVC~\cite{FGVC} for fine-grained classification; SUN397~\cite{SUN397} for scene recognition; and DTD~\cite{DTD} for texture recognition. 
We compare with TeCoA~\cite{TeCoA}, PMG~\cite{PMG}, and LAAT~\cite{LAAT} under the zero-shot setting. All methods are fine-tuned on Tiny-ImageNet~\cite{ImageNet} with matched training settings for fair comparison. We evaluate zero-shot robustness under PGD~\cite{AT}, CW~\cite{CW}, AutoAttack~\cite{AA}, and BIM~\cite{BIM}.

\noindent{\textbf{Implementation Details.}}
We use the CLIP-B/32 architecture with the prompt ``This is a photo of a \{\}". The model is fine-tuned on Tiny-ImageNet for 10 epochs (batch size 512, learning rate $10^{-4}$) using an SGD optimizer (momentum 0.9). Adversarial training employs a 2-step PGD attack ($l_{\infty}, \varepsilon=1/255$, step size 1/255). The source code is included in the \textit{supplementary materials}.

\subsection{Result and Analysis}
\label{exp:results-analysis}
\subsubsection{Compared with SOTA Methods.}
\label{exp:results}
 
Tab.~\ref{tab:zero-shot-tiny} shows the zero-shot robust accuracy under PGD attack and zero-shot clean accuracy of SOTA methods and our proposed approach TIMA under the $\ell_{\infty}$ setting, with the maximum adversarial perturbation set to $\varepsilon=1/255, 8/255 $. Regarding the zero-shot robust accuracy, under small bound attacks ($\varepsilon=1/255$), zero-shot robust accuracy of TIMA exceeds TeCoA by 4.69\%, PMG by 2.80\%, and LAAT by 6.64\%. When facing a large attack ($\varepsilon = 8/255$), zero-shot robust accuracy of our proposed TIMA exceeds TeCoA by 13.30\%, PMG by 16.22\%, and LAAT by 9.95\%. Our proposed method shows a greater performance improvement over other methods under an 8/255 perturbation bound attack compared to the performance improvement observed under a 1/255 perturbation bound attack. Besides zero-shot adversarial robustness, on zero-shot generalization (clean accuracy), TIMA improves by 5.50\% compared with TeCoA, 3.08\% compared with PMG, and 11.53\% compared with LAAT. From Tab.~\ref{tab:zero-shot-tiny}, we conclude that our method also has a favorable perfect performance both in zero-shot adversarial robustness and zero-shot generalization. For results on adversarial fine-tuning on ImageNet, please refer to \textit{Appendix}. In addition to clean and robust accuracy, we report training cost and inference efficiency to assess overall computation in the \textit{Appendix}.

\subsubsection{Adversarial Evaluation Across Multiple Attacks.}
As illustrated in Fig.~\ref{fig:CW-AA-BIM}, we conducted a comprehensive robustness evaluation across a continuous spectrum of perturbation magnitudes (from $\varepsilon=1/255$ to $8/255$) under BIM attack.  The results consistently demonstrate that our proposed TIMA method (green triangles) maintains a significant and stable performance advantage over baseline methods like TeCoA and PMG across all tested datasets.  Notably, this performance gap widens dramatically as the perturbation bound $\varepsilon$ increases.  While the robust accuracy of baseline methods drops sharply and tends to collapse at larger perturbations (especially $\varepsilon \ge 4/255$), TIMA exhibits a much more gradual performance drop, showcasing its superior robustness against stronger attacks.  This superior performance directly validates the core motivation of our work: by identifying and addressing the stable logit margin offset between small and large perturbations, our novel ASAM and SC-MHE module explicitly calibrates decision boundaries during training. This enables the model to effectively generalize the robustness learned from small perturbations to unseen, larger perturbations at inference time, thus achieving a more comprehensive and reliable defense across a wide range of attack perturbation magnitudes.
For results on more datasets under CW, BIM, and AutoAttack, please refer to {\textit{Appendix}}.

\begin{figure}[t!]
    \centering
    \includegraphics[width=\linewidth]{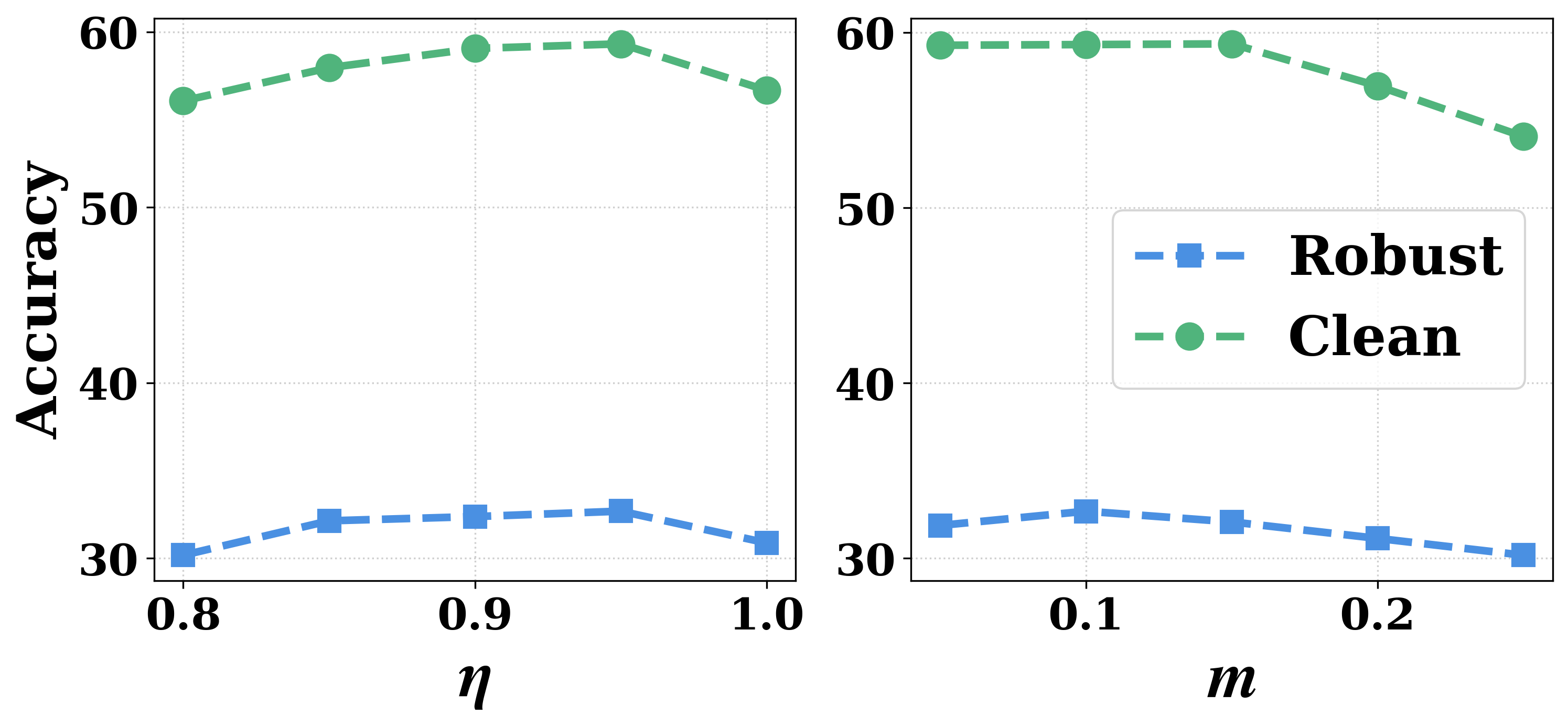}
    \caption{ASAM Sensitivity to Hyperparameters $m$ and $\eta$. 
}
    \label{fig:hypers}
\end{figure}

\subsubsection{Visualization Analysis.}
The t-SNE visualizations in Fig.~\ref{fig:tSNE} demonstrate the superiority of our method under strong adversarial perturbations (CIFAR-10, $\varepsilon=8/255$). Baseline methods, particularly TeCoA, suffer from severe {feature collapse}, with embeddings from nearly all classes tangled together. While PMG and LAAT form some discernible clusters, they still exhibit significant {inter-class overlap} and {intra-class dispersion}. In contrast, TIMA learns a much cleaner feature space, maintaining both {tight intra-class clusters} and {clear inter-class boundaries}. This provides strong evidence of TIMA's superior ability to preserve discriminative features and semantic structure, thereby enhancing robustness.

\subsubsection{Adversarial Robustness without Adversarial Training.} 
A key finding, detailed in Tab.~\ref{tab:zero-shot-clean-training}, is the emergent adversarial robustness achieved by TIMA even when trained exclusively on clean data. Without any exposure to adversarial examples during fine-tuning, TIMA achieves an average robust accuracy of 11.72\% against $\ell_{\infty}$ PGD attacks ($\varepsilon=1/255$), a substantial +10.85\% improvement over the TeCoA baseline. This phenomenon is not accidental but an emergent property of our framework's design. The ASAM module, by enforcing semantic-aware margins even on clean samples, acts as an implicit robustness regularizer, creating more resilient decision boundaries. Concurrently, the SC-MHE module optimizes the geometry of the class prototype space by uniformly separating text embeddings while preserving semantic consistency. The synergy between these two components fundamentally enhances the model's intrinsic robustness, demonstrating that TIMA improves the model's core structure rather than merely learning to defend specific attack patterns.

\subsubsection{\textcolor{black}{Guidelines and Sensitivity Analysis of $\eta$ and $m$ in ASAM.}}

We conduct a sensitivity analysis of the ASAM hyperparameters $\eta$ and $m$ (see Fig.~\ref{fig:hypers}). As shown in the figure, both clean and robust accuracies remain stable when $m \in [0.05, 0.15]$ and $\eta \in [0.85, 0.95]$, and our default setting $(m = 0.10,\ \eta = 0.95)$ lies in this stable region. Detailed analysis and selection strategies are provided in \textit{Appendix}.

\subsection{Ablation Study}
\label{exp:ablation}

Tab.~\ref{tab:ablation-SAM-IAT} presents the results of an ablation study on two core parts, ASAM and SC-MHE, showing the zero-shot clean accuracy and zero-shot robust accuracy across multiple test datasets under a large perturbation bound ($\varepsilon = 8/255$).  Besides these results, Fig.~\ref{fig:CLIP-LAAT-ours-text-text} (c) shows the relationship between clean image embeddings and tuned text embeddings, which presents less semantic inconsistency than those of LAAT. 


\begin{table}[t!]
\centering
\small 
\tabcolsep=1pt
\extrarowheight=1.7pt
\begin{tabular}{lc|cccc}
\toprule
\textbf{Methods} & \hspace{1.2em}$\varepsilon$\hspace{1.2em} & \textbf{CIFAR10} & \textbf{CIFAR100} & \textbf{STL10} & \textbf{ImageNet} 
\\
\midrule
TeCoA & 0 & 76.10	   & 46.00	    & 91.39	& 43.87  
\\
 w/ ASAM & 0 & 80.41 & 51.87 & 92.40 & 42.75
\\
 w/ SC-MHE & 0 &  76.74 &45.92 &91.94 &44.50
\\
\cdashline{1-6}  
\Tstrut \Bstrut
TIMA (Ours) & 0 &  	 \textbf{84.23}	   & \textbf{57.90}	    & \textbf{93.61}	& \textbf{46.55} 	  \\
\midrule
TeCoA & $\nicefrac{8}{255}$ & 7.06	   & 3.95	    & 40.93	 & 1.85 
\\
w/ ASAM & $\nicefrac{8}{255}$ & {31.56} & \textbf{17.44} & 67.11 & \textbf{12.77}
\\
 w/ SC-MHE & $\nicefrac{8}{255}$ & 23.62 & 14.08 & 48.97  & 5.64 
\\
 \cdashline{1-6}  
\Tstrut \Bstrut
TIMA (Ours) & $\nicefrac{8}{255}$ &  \textbf{32.45}	   & {14.38}	   & \textbf{73.84}	 & {10.21}  	 
\\
\bottomrule
\end{tabular}
\caption{Ablation Study of the proposed ASAM and SC-MHE module. ``w/" donates TeCoA with different modules.}
\label{tab:ablation-SAM-IAT}
\end{table}
\section{Conclusion}

In this work, we propose the Text-Image Mutual Awareness (TIMA) framework to balance zero-shot robustness and generalization. Its TAI module uses an adaptive margin (ASAM) to enhance robustness, while its IAT module employs semantic-consistent energy minimization (SC-MHE) to preserve generalization. We show that co-optimizing logit margins and semantic alignments is crucial for building robust, generalizable models.


\section*{Acknowledgments}
This work was supported in part by the National Natural Science Foundation of China (No. 62471420), GuangDong Basic and Applied Basic Research Foundation (2025A1515012296), and 2025 Tencent AI Lab Rhino-Bird Program. Hei Victor Cheng is supported in part by the Aarhus Universitets Forskningsfond project number AUFF 39001 and the NordForsk Nordic University Cooperation on Edge Intelligence (Grant No. 168043).

\bibliography{aaai2026}

\end{document}